\documentclass[letterpaper, 10 pt, conference]{ieeeconf}  %

\IEEEoverridecommandlockouts                              %

\usepackage[utf8]{inputenc}
\usepackage[T1]{fontenc}
\usepackage{hyperref}
\usepackage{url}
\usepackage{booktabs}
\usepackage{amsmath,amssymb,amsfonts}
\usepackage{nicefrac}
\usepackage{microtype}
\usepackage{xcolor}
\usepackage{graphicx}
\usepackage{algorithm}
\usepackage{algpseudocode}
\usepackage{multirow}
\usepackage{morefloats}
\algrenewcommand\algorithmiccomment[1]{\hspace{0.5em}\(\triangleright\) #1}

\usepackage[capitalise,nameinlink]{cleveref}

\crefname{figure}{Fig.}{Figs.}
\Crefname{figure}{Fig.}{Figs.}

\crefname{section}{Sec.}{Secs.}
\Crefname{section}{Sec.}{Secs.}

\crefname{equation}{Eq.}{Eqs.}
\Crefname{equation}{Eq.}{Eqs.}

\crefname{table}{Table}{Tables}
\Crefname{table}{Table}{Tables}

\crefname{algorithm}{Alg.}{Algs.}
\Crefname{algorithm}{Alg.}{Algs.}

\usepackage{caption}

\usepackage[
    backend=biber,
    style=numeric,
    citestyle=numeric,
    sorting=none,
    sortcites=true,
    natbib=true
]{biblatex}

\title{\LARGE \bf
Closed-Loop Adaptive Sampling on a Generative Ocean Model
}

\author{Shayesteh Hafezi$^{1}$ and Alan Papalia$^{1}$%
\thanks{$^{1}${\tt\small \{shafezi, apapalia\}@umich.edu}, 
University of Michigan, Ann Arbor, MI 48109, USA.}%
\thanks{$^{2}$Code, model weights, and per-run evaluation records are available at
\url{https://github.com/UMich-RobotExploration/closed-loop-sampling}.}%
}

\begin{document}

\maketitle
\thispagestyle{empty}
\pagestyle{empty}

\begin{abstract}
Adaptive sampling can substantially improve the value of sparse ocean
observations by directing mobile robots toward measurements that are most
informative for an evolving environment.
Recent learned generative ocean models can produce high-resolution forecasts and
ensembles at substantially lower computational cost than traditional numerical
models, creating an opportunity to use rich predictive forecasts directly in
closed-loop robotic sampling.
However, these models do not natively provide the explicit uncertainty
representations assumed by classical informative-planning methods, and their
forecasts often evolve in learned latent spaces rather than directly in physical
ocean variables. 
We develop a multi-robot closed-loop adaptive sampling framework that uses a
single learned generative model for forecasting, uncertainty estimation,
planning, and measurement assimilation.
The model forecasts an ensemble in latent space, decodes it to physical ocean
states, and uses ensemble statistics to construct A- or D-optimal sampling
objectives.
We select sampling locations subject to robot reachability under predicted
currents. The collected measurements are assimilated back into the latent state
using observation-conditioned diffusion sampling.
Across 120 test scenarios A- and D-optimal sampling generally outperforms static, random, and systematic-coverage baselines 
while adding little planning overhead. 
Our experiments uncover several scenarios in which model uncertainty diverges from forecast error, motivating future work in uncertainty quantification.$^{2}$

\end{abstract}

\section{Introduction}
\label{sec:intro}

The ocean plays a central role in weather, climate, ecosystems, and maritime
activity \cite{lenton2019climate,rayner2019ocean}. However, it remains difficult to
observe because it is vast, dynamic, and sparsely measured \cite{garwood2026recent}.
Mobile robots can help close these data gaps via adaptive sampling, directing
measurements toward locations and times that are expected to be most informative
\cite{lermusiaux2007adaptive,popovic2024aipp}.

Effective adaptive sampling depends on a predictive belief over the evolving
ocean.
This belief must forecast the quantities of interest, represent uncertainty well
enough to identify informative measurements, and provide the flow fields that
affect robot motion and reachability \cite{lermusiaux2017optimal,lermusiaux2024mastr,mule2024reachability}.
Existing approaches typically rely on physics-based numerical forecasts and
ensembles \cite{lermusiaux2007adaptive,lermusiaux2024mastr,moore2019synthesis} or compact statistical models such as Gaussian processes
\cite{hollinger2014sampling,fossum2020compact}.
Numerical models provide rich predictions of ocean dynamics, but repeated
forecasting, ensemble generation, and data assimilation can be computationally
expensive \cite{moore2019synthesis}.
Compact statistical models make uncertainty-aware planning more tractable, but
at the cost of a simplified representation of the environment.

\begin{figure}[tb]
\centering
\includegraphics[width=0.99\linewidth]{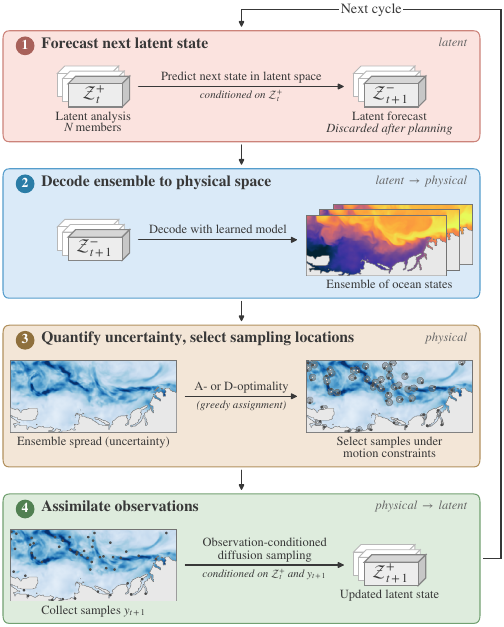}
\caption{\textbf{Method overview.}
At each timestep, our method
(1) forecasts an ensemble of ocean states in a learned latent space,
(2) decodes the ensemble to a physical representation of ocean state,
(3) uses ensemble spread to estimate uncertainty and assigns robots to informative, reachable sampling locations using A- or D-optimality with greedy assignment, and
(4) assimilates the resulting observations via observation-conditioned diffusion sampling to update the latent state for the next cycle.
}

\label{fig:sys_fig}
\end{figure}

Recent learned ocean forecasting models are shifting the tradeoff between model
fidelity and computational demand \cite{holmberg2026njord,huang2026fuxions}.
They can produce high-resolution forecasts with substantially lower rollout cost
than traditional numerical models.
Generative variants of learned forecasts can efficiently produce ensembles of
plausible future ocean states \cite{zhuang2025ladcast}.
This creates an opportunity to perform adaptive sampling directly with rich
predictive forecasts rather than simplified surrogates.

However, these learned forecasting models introduce new challenges for adaptive sampling.
Their predictive uncertainty is generally available only implicitly through
generated samples rather than through an explicit probabilistic belief.
Moreover, the forecast often evolves in a learned latent space, while sensing
and planning objectives are defined over physical ocean variables that can only
be accessed through a learned decoder.
The improved efficiency and fidelity of learned ocean forecasts, in combination with the
stated challenges, motivates the development of adaptive sampling frameworks that are
designed specifically around these generative models.

\textbf{This paper.}
We develop a multi-robot closed-loop adaptive sampling framework that uses a
single learned generative model for forecasting, informative sampling, and
measurement assimilation.
Ocean states are represented in a learned latent space, where the model
generates an ensemble of possible future states.
We use a learned decoder to map this ensemble to physical ocean variables and
use forecast disagreement to estimate sampling value.
Sampling locations are selected using A- or D-optimality, subject to each
robot's reachable set under the predicted currents, with greedy assignment
across the team.
The resulting measurements are then assimilated using observation-conditioned
diffusion sampling to update the latent ocean-state estimate.

We summarize our contributions as follows:
\begin{itemize}

\item The first closed-loop robotic sampling framework where a learned generative forecast serves as the predictive belief for forecasting, planning, and assimilation.

\item We enable uncertainty-aware multi-robot placement with A- and D-optimal objectives, showing that adaptive sampling remains practical with high-dimensional spatiotemporal generative forecasts.

\item Across 120 test scenarios, we show when generative uncertainty improves
adaptive sampling and identify real-world failure modes when ensemble spread is
poorly aligned with forecast error.
\end{itemize}

\section{Related Work}
\label{sec:related}

\textbf{Adaptive sampling in the ocean.}
Adaptive sampling has been studied with several kinds of environmental
beliefs. Gaussian-process and related stochastic field models have been
used onboard AUVs to guide sampling of fronts, phytoplankton, and other
features
\citep{fossum2019phytoplankton,fossum2021excursion,berget2023dynamic}.
These models provide useful estimates of the local field, but they do not
generally provide a full forecast of the evolving ocean state that the
robot will encounter. Numerical forecast ensembles do provide such a
predictive belief, and ensemble uncertainty has long been used to value
candidate observations
\citep{bishop2001etkf,majumdar2002etkf2}.
In the ocean, the forecast can also affect where observations are
possible, since predicted currents constrain the locations a robot can
reach
\citep{lolla2014timeoptimal,subramani2018stochastic,kulkarni20203dpath}.
Several ocean experiments have combined forecast-aware motion with
informative sampling
\citep{wang2009acoustically,lermusiaux2017optimal,lermusiaux2024mastr,
mirabito2025realtime}, and others have closed the loop by assimilating
robot observations and using the updated forecast to retask the platform
\citep{ford2022adaptive,halvorsen2026closedloop}.
These works establish closed-loop ocean sampling using statistical or numerical forecast models. Our work instead uses a learned generative forecast as the belief that guides where robots sample.

\textbf{Generative geophysical forecasting.}
Learned geophysical models have demonstrated orders-of-magnitude faster inference than state-of-the-art numerical prediction systems while maintaining competitive forecast skill
\citep{bi2023accurate,lam2023learning,price2025gencast}.
Analogous learned ocean models now provide high-resolution, eddy-resolving forecasts
\citep{chattopadhyay2024oceannet,wang2024xihe,cui2025wenhai,elaouni2025glonet,dheeshjith2025samudra}.
Most of these ocean models are deterministic, however, and do not provide predictive uncertainty estimates.
Generative forecasters address this limitation by producing ensembles of plausible future states; such models have achieved competitive probabilistic forecast skill in the atmosphere
\citep{price2025gencast,zhuang2025ladcast} and ocean \citep{holmberg2026njord,huang2026fuxions}.
Rather than only rolling a generative forecast forward from its initial condition, where errors can accumulate over time, we use its forecast to decide where robots sample and then update the belief with the observations they collect before forecasting again.

\textbf{Generative data assimilation and sensor placement.}
Recent work has studied how observations can be incorporated into learned
generative models by treating data assimilation as posterior inference under a
generative prior
\citep{rozet2023score,huang2024diffda}.
This has also provided updates for latent generative representations
\citep{andry2025appa} and to incorporate sparse ocean observations
\citep{martin2025genda,asefi2025generative}.
Separate recent work studies static sensor placement when the
environmental prior is represented by a generative model
\citep{rowbottom2026christoffel,chapron2026oceanplacement}.
These different works address important pieces of the problem, but typically separate forecasting, data
assimilation, and sensor placement.
Our work differs by showing how to use a generative forecast for all of these aspects within a single closed loop, and in the case of mobile sampling platforms.
\section{Closed-Loop Sampling}
\label{sec:method}

\textbf{Overview.}
We consider a fleet of $n$ robots that repeatedly selects
sampling locations in an evolving ocean field, subject to motion
constraints. Let $t=0,\ldots,T$ index observation times separated
by $\Delta t$. The robots maintain an $N$-member ensemble
representing their belief over the ocean state. After initialization,
the reference state is unavailable and the robots receive only the
point measurements collected by the fleet.

A learned autoencoder $(\phi_E,\phi_D)$ maps the multivariable
ocean state to and from a compressed latent representation.
We denote the forecast and analysis ensembles by
$\mathcal Z_t^-=\{\mathbf z_t^{(i),-}\}_{i=1}^N$ and
$\mathcal Z_t^+=\{\mathbf z_t^{(i),+}\}_{i=1}^N$,
respectively.

Starting from $\mathcal Z_t^+$, the generative forecast produces
the next forecast ensemble $\mathcal Z_{t+1}^-$. We decode this
ensemble to evaluate candidate sampling locations and determine
robot reachability from the predicted surface currents. After
the robots move and collect observations, observation-guided
generation produces the next analysis ensemble
$\mathcal Z_{t+1}^+$. One closed-loop iteration is therefore
{\setlength{\abovedisplayskip}{4pt}
\setlength{\belowdisplayskip}{4pt}
\begin{equation*}
\boxed{
\mathcal Z_t^{+}
\xrightarrow{\ \text{\tiny forecast}\ }
\mathcal Z_{t+1}^{-}
\xrightarrow{\ \text{\tiny select}\ }
\mathbf p_{t+1}
\xrightarrow{\ \text{\tiny observe}\ }
\mathbf y_{t+1}
\xrightarrow{\ \text{\tiny update}\ }
\mathcal Z_{t+1}^{+},
}
\end{equation*}
}
where $\mathbf p_{t+1}=\{\mathbf p_{r,t+1}\}_{r=1}^n$
denotes the fleet's next sampling locations. Assimilation generates
new samples conditioned on the previous analysis and the collected
observations; it does not directly correct the forecast members.
Algorithm~\ref{alg:cycle} summarizes this process.

At $t=0$, we initialize all ensemble members by encoding the
reference state with $\phi_E$. The initial ensemble has zero
spread, and its decoded error is the autoencoder reconstruction
error. Spread first appears through stochastic forecasting, and
the encoder is used only for initialization. Across all policies
we compare, only the sampling-location selection rule changes;
forecasting, motion, observation, and assimilation are otherwise
shared.

\begin{figure*}[h!]
\centering
\includegraphics[width=0.99\textwidth]{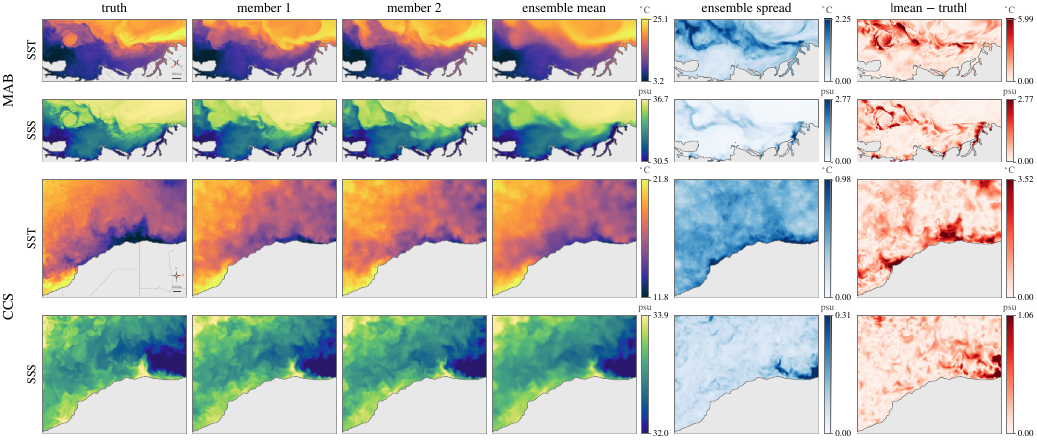}
\caption{\textbf{Forecast ensembles and error in the two regions.}
Illustrative forecasts at 240\,h for MAB (2022-04-09, top two rows)
and CCS (2016-07-27, bottom two rows), with SST above SSS in each
region. Columns show the reference state, two ensemble members,
ensemble mean, ensemble spread, and absolute error of the mean. On
MAB, error concentrates around the shelf-slope front and a warm-core
eddy, and the spread highlights many of the same structures. CCS errors
are finer-grained and more broadly distributed; spread--error alignment
is weaker on average across its temperature ICs, although the example
shown here is relatively well aligned ($r=0.44$).}
\label{fig:show_both}
\vspace{-2mm}
\end{figure*}

\begin{algorithm}[t]
\caption{One cycle of closed-loop sampling}
\label{alg:cycle}
\footnotesize
\begin{algorithmic}[1]
\Require analysis latents $\{\mathbf z_t^{(i),+}\}_{i=1}^N$, robot positions
$\{\mathbf p_{r,t}\}_{r=1}^n$, observation-noise variance $R$, guidance
scale $\lambda_g$
\For{$i=1,\dots,N$}
  \State sample $\mathbf z_{t+1}^{(i),-}$ from the forecast model
  given $\mathbf z_t^{(i),+}$
  \State $\mathbf x_{t+1}^{(i),-}
  \gets\phi_D(\mathbf z_{t+1}^{(i),-})$;
  $\mathbf s^{(i)}\gets
  P_{\mathrm{tar}}\mathbf x_{t+1}^{(i),-}$
\EndFor
\State $\bar{\mathbf u}\gets\tfrac12\big(
\mathrm{mean}_i\,\mathbf u(\phi_D(\mathbf z_t^{(i),+}))
+\mathrm{mean}_i\,\mathbf u(\mathbf x_{t+1}^{(i),-})\big)$
\For{$r=1,\dots,n$}
  \State $\mathcal R_r\gets$ wet cells within $v\Delta t$ of
  $\mathbf p_{r,t}$ advected by $\bar{\mathbf u}$
\EndFor
\State $X\gets[\mathbf s^{(1)}-\bar{\mathbf s},\dots,
\mathbf s^{(N)}-\bar{\mathbf s}]$
\State $\mathbf p_{t+1}\gets
\textsc{SelectLocations}(X,\{\mathcal R_r\},R)$ \hspace{-0.4em}
\Comment{greedy A-/D-optimal assignment}
\State $\mathbf y_{t+1}\gets
H_{\mathbf p_{t+1}}(\mathbf s^\star_{t+1})+\boldsymbol\eta$,
$\quad\boldsymbol\eta\sim\mathcal N(0,RI_n)$
\For{$i=1,\dots,N$}
  \State draw $\boldsymbol\epsilon\sim\mathcal N(0,I)$;
  $\mathbf w\gets\sigma_{\max}\boldsymbol\epsilon$
  \For{diffusion step $k$}
    \State $\widehat{\mathbf z}_0\gets
    D_\theta(\mathbf w;\mathbf z_t^{(i),+},\sigma_k)$
    \State $L\gets R^{-1}\lVert
    H_{\mathbf p_{t+1}}(P_{\mathrm{tar}}\phi_D(\widehat{\mathbf z}_0))
    -\mathbf y_{t+1}\rVert_2^2$
    \State $\mathbf w\gets
    \textsc{SamplerStep}(\mathbf w)-\lambda_g\nabla_{\mathbf w}L$
  \EndFor
  \State $\mathbf z_{t+1}^{(i),+}\gets\mathbf w$
\EndFor
\State \Return $\{\mathbf z_{t+1}^{(i),+}\}_{i=1}^N$, $\mathbf p_{t+1}$
\end{algorithmic}
\end{algorithm}

\textbf{Forecast.}
We adapt the latent diffusion forecasting model LaDCast
\cite{zhuang2025ladcast}.
Its autoencoder has $\approx256$\,M parameters and compresses the ocean state by
a factor of eight in each horizontal dimension.
A $375$\,M-parameter denoiser uses the preconditioning and sampling procedure of
\cite{karras2022edm}.

For each ensemble member $i$, we draw an independent Gaussian noise
sample in latent space and use the diffusion model, conditioned
on the member's previous analysis latent $\mathbf z_t^{(i),+}$,
to generate the next-state forecast $\mathbf z_{t+1}^{(i),-}$.
Each model call predicts four
successive states spaced $\Delta t$ apart; we retain only the
first to advance the ensemble by one observation interval.

Decoding each forecast latent gives the multivariable physical
state
\(
\mathbf x_{t+1}^{(i),-}
=\phi_D(\mathbf z_{t+1}^{(i),-}).
\)
A projection $P_{\mathrm{tar}}$ extracts the variable and depth
to be sampled,
\(
\mathbf s_{t+1}^{(i),-}
=P_{\mathrm{tar}}\mathbf x_{t+1}^{(i),-},
\)
while the decoded surface currents determine robot reachability.

We adapt LaDCast's global-grid formulation to each regional
ocean domain using a binary wet-mask input channel, wet-masked
loss, replicate boundary padding, and latent-grid positional
indices. We omit seasonal conditioning and train a separate
model for each region (\cref{sec:setup}).

\textbf{Sampling.}
We first determine the locations each robot can reach using the
predicted surface currents. Because the forecast produces an ensemble
of current fields, we use their mean for planning. To approximate
the current over the observation interval, we average the
ensemble-mean surface currents decoded from the analysis ensemble
at time $t$ and the forecast ensemble at time $t+1$.

Robot $r$ is advected by this current over $\Delta t$ and can
travel an additional distance $v\Delta t$, where $v$ is its
propulsion speed. Its reachable set $\mathcal R_{r,t+1}$
contains the wet cells within this distance of the advected
position.

We approximate forecast uncertainty using the empirical
covariance $\widehat C=(N-1)^{-1}XX^\top$, where
\[
X=[\mathbf s^{(1)}-\bar{\mathbf s},\ldots,
\mathbf s^{(N)}-\bar{\mathbf s}]
\in\mathbb R^{d\times N}
\]
contains the deviations of each forecast member from the
ensemble mean. Here, $\mathbf s^{(i)}=\mathbf s_{t+1}^{(i),-}$
is the predicted target field over $d$ wet cells,
$\bar{\mathbf s}$ is the ensemble mean, and $x_j^\top$
denotes row $j$ of $X$. We never form $\widehat C$ explicitly.

We use two covariance-based acquisition criteria to score candidate
sampling locations. D-optimality measures the information gained
from an observation under a Gaussian belief approximation, whereas
A-optimality measures the corresponding reduction in total field
variance. For observation-noise variance
$R=\sigma_{\mathrm{obs}}^2$, their single-observation acquisition
scores are
{\setlength{\abovedisplayskip}{6pt}
\setlength{\belowdisplayskip}{6pt}
\begin{equation}
u_{\mathrm D}(j)
=
\tfrac12\log\left(1+\frac{\widehat C_{jj}}{R}\right),
\label{eq:dopt}
\end{equation}
}
{\setlength{\abovedisplayskip}{6pt}
\setlength{\belowdisplayskip}{6pt}
\begin{equation}
u_{\mathrm A}(j)
=
\frac{\|\widehat C_{:,j}\|_2^2}
     {\widehat C_{jj}+R}
=
\frac{x_j^\top Gx_j}
     {(N-1)^2(\widehat C_{jj}+R)},
\label{eq:aopt}
\end{equation}
}
where $G=X^\top X$. Both scores are evaluated using the current
covariance approximation, with larger values indicating more
informative locations.

For a Gaussian belief, $u_{\mathrm D}$ is the information gain
from observing cell $j$, while $u_{\mathrm A}$ is the reduction
in covariance trace under a rank-one Kalman covariance update
\citep{bishop2001etkf,hamill2002adaptive,krause2008sensor}.
D-optimality therefore favors cells with large local variance,
whereas A-optimality also accounts for covariance with the rest
of the field:
\[
\|\widehat C_{:,j}\|_2^2
=
\widehat C_{jj}^2+
\sum_{k\neq j}\widehat C_{kj}^2.
\]
Both criteria use only the empirical covariance and can therefore
be applied to ensembles from the non-Gaussian generative forecast.
After constructing $G$, the Gram-matrix form in
Eq.~\eqref{eq:aopt} evaluates the A-optimal score in
$O(N^2)$ per candidate rather than $O(dN)$.

We assign sampling locations by greedy selection. At each step,
we choose the feasible robot--cell pair with the largest current
acquisition score. After selecting a location, we apply a
hypothetical serial ensemble square-root update to the anomalies
\citep{whitaker2002ensrf,tippett2003esrf}. This planning update
requires no measurement: it accounts for the uncertainty reduction
that would result from observing the selected cell, thereby
reducing the marginal value of redundant observations for the
remaining robots.

Each robot may be assigned at most one cell, and each cell may
be assigned to at most one robot. These constraints define the
intersection of two partition matroids over feasible robot--cell
pairs. Because the Gaussian D-optimal information-gain objective
is monotone and submodular, greedy selection achieves a
$\nicefrac13$ approximation guarantee
\citep{fisher1978analysis,krause2008sensor}.
We use the same procedure for A-optimality, but this guarantee
does not generally apply because the variance-reduction objective
is not submodular in general \citep{krause2008sensor}.

\textbf{Assimilation.}
The observations collected by the robots are used to update the latent
ocean state before the next forecast. We model them as
$\mathbf y_{t+1}
=H_{\mathbf p_{t+1}}(\mathbf s_{t+1}^{\star})+\boldsymbol\eta$,
with $\boldsymbol\eta\sim\mathcal N(0,RI)$, where
$H_{\mathbf p}$ samples the target field at locations $\mathbf p$.
Because generation takes place in latent space while the observations
are defined in physical space, we condition the latent generation using
likelihood guidance adapted from diffusion posterior sampling
\citep{chung2023dps}. At each reverse-diffusion step, the denoiser
estimates the clean latent associated with the current noisy state. We
decode this estimate, extract the target field, compare it with the observations, and use the resulting residual to guide the
latent trajectory.

For ensemble member $i$, let $\mathbf w_k^{(i)}$ denote the current noisy
latent at diffusion level $\sigma_k$. Conditioned on the previous
analysis latent $\mathbf z_t^{(i),+}$, the denoiser estimates the clean
latent $\widehat{\mathbf z}_{0,k}^{(i)} = D_\theta(\mathbf w_k^{(i)};
\mathbf z_t^{(i),+}, \sigma_k)$, whose decoded target field is
$\widehat{\mathbf s}_{0,k}^{(i)} = P_{\mathrm{tar}}\phi_D
(\widehat{\mathbf z}_{0,k}^{(i)})$.
The decoded estimate can then be compared directly with the observations
through $L_k^{(i)} = R^{-1}\lVert H_{\mathbf p_{t+1}}
(\widehat{\mathbf s}_{0,k}^{(i)}) - \mathbf y_{t+1}\rVert_2^2$, with
$\mathbf g_k^{(i)} = \nabla_{\mathbf w_k^{(i)}}L_k^{(i)}$.
The gradient is propagated through the frozen denoiser and decoder to the
current noisy latent. The direction $-\mathbf g_k^{(i)}$ therefore moves
the latent trajectory toward states whose decoded target field agrees
more closely with the observations.

At each diffusion level, we first advance the latent with the same
sampler used for forecasting, producing
$\widetilde{\mathbf w}_{k-1}^{(i)}$, and then apply the observation
guidance,
\(
\mathbf w_{k-1}^{(i)}
=
\widetilde{\mathbf w}_{k-1}^{(i)}
-
\lambda_g\mathbf g_k^{(i)} .
\)
This procedure is repeated through the reverse-diffusion trajectory.
Each analysis member starts from a fresh Gaussian draw and is guided
independently using the same observations. The resulting member is
therefore a new observation-conditioned sample rather than a correction
of its corresponding forecast member; a draw of the next
latent state given both the previous analysis and the new observations.
 
Although the observations constrain only the target variable selected by
$P_{\mathrm{tar}}$, the guidance acts on the full latent state. Other
components of the decoded ocean state can therefore change as well,
including the surface currents used for reachability at the next
observation time.

\section{Experimental Setup}
\label{sec:setup}

We evaluate our methodology for both A-optimal and D-optimal sampling criteria against
three baselines on 1,800 sequences. Each sequence is 10 days total, split into 6
hour increments.  The sequences are taken from reanalysis products of two
different ocean regions.

\textbf{Regions and data.}
We use two ocean region datasets with different geometries and dynamics. The
first is a Mid-Atlantic Bight (MAB) reanalysis
\citep{wilkin2022doppio_reanalysis}, on a rotated curvilinear grid with
near-square $\sim$7\,km cells and $19{,}150$ wet cells. The second is a
California Current System (CCS) reanalysis \citep{neveu2016ccsra}, on a
$0.1^{\circ}$ grid with $8.7 \times 11.1$\,km cells and $21{,}240$ wet
cells.

We train one forecast model per region. The models use the same architecture, parameter counts, and sampler. Each autoencoder maps an 84-channel field, comprising velocity, temperature and salinity on 20 terrain-following levels, sea surface height, the two depth-averaged velocities, and a binary wet mask, to a latent of 84 channels at one eighth the horizontal resolution in each direction. The models differ only in latent tensor shape ($84\times13\times30$ for MAB and $84\times22\times23$ for CCS), where the first dimension is the number of latent channels.
The MAB model is trained on 2007--2020 and
validated on 2021; the CCS model is trained on 1980--2014 and validated
on 2015. The years 2022-2024 and 2016--2019 are held out as test data
for the MAB and CCS system, respectively.

\textbf{Sampling policies.}
We compare five policies: \emph{Static}, \emph{Random}, \emph{Lawnmower}, \emph{A-optimal}, and
\emph{D-optimal}.

\emph{A-optimal} and \emph{D-optimal}, defined in \cref{sec:method}, evaluate
reachable cells using the acquisition criteria in
\cref{eq:dopt,eq:aopt} and assign robots greedily.
Both are myopic, selecting locations only for the next observation time.
\emph{Static} samples the same locations at every cycle, while \emph{Random} draws
each robot's next location uniformly from its reachable set.
\emph{Lawnmower} follows a survey route fixed before the run. The wet domain is
divided among the robots, and a back-and-forth route is planned for each
assigned region \cite{choset1998coverage,rekleitis2008coverage}. 

For \emph{Static}, \emph{Random}, \emph{A-optimal}, and \emph{D-optimal}, the $n$ robots start on
an approximately uniform lattice. Any position that falls on land is
moved to the nearest wet cell. \emph{Static} remains at these locations,
whereas the three mobile strategies move from them. \emph{Lawnmower} instead
starts each robot at the beginning of its planned route.

\textbf{Experiments.}
Each scenario is run twice, once for sea-surface temperature (SST) and once for
sea-surface salinity (SSS). 
The experiments sweep over fleets sizes of $n\in\{20,50,200\}$ robots. In all experiments, forecast ensembles of size $N=25$ are used.

Observations are collected every $\Delta t=6\,\mathrm{h}$ for $T=40$ cycles,
giving a 240\,h evaluation horizon. 
The robot propulsion speed is
$v=1.5\,\mathrm{m/s}$, so, without currents, each robot could travel up to
$v\Delta t=32.4$\,km for each observation interval.
Temperature observations use $\sigma_{\mathrm{obs}}=0.2\,^{\circ}$C.
For salinity, an equivalent sensor on MAB would have
$\sigma_{\mathrm{obs}}=0.143$\,psu (estimated from 200 validation
states); we use this value, fixed across both regions.

The guidance scale $\lambda_g$ is selected for each region, variable, and fleet size from a sweep over fifteen ICs from the validation year using
D-optimal observation placement, then fixed for evaluation and shared
across all five sampling strategies. The selected values range from
$10^{-4}$ to $2.2\times10^{-2}$.

The evaluation comprises 2 regions $\times$ 2 variables $\times$ 3 fleet sizes $\times$ 10 ICs, giving 120 test cases. Each test case is evaluated with 5 sampling strategies $\times$ 3 seeds, for a total of 1,800 runs of 40 cycles each, with seeds paired across
strategies via common random numbers \citep{glasserman1992crn}.

{
\setlength{\tabcolsep}{3pt}
\renewcommand{\arraystretch}{0.9}
\begin{table}[t]
\centering\footnotesize
\setlength{\tabcolsep}{4.5pt}
\caption{RMSE of the analysis-ensemble mean ($^{\circ}$C for SST, psu for
SSS) over the final eight cycles, averaged over ten held-out dates and
three paired seeds. Bold: best per row, from unrounded scores.
$^\dagger$: tied with the best (not reliably separated by the paired test and rerun floor, Sec.~\ref{sec:setup})}
\label{tab:summary}
\begin{tabular}{ll ccccc}
\toprule
 & $n$ & \emph{Static} & \emph{Random} & \emph{Lawnmower} & \emph{D-optimal} & \emph{A-optimal} \\
\midrule
\multirow{3}{*}{\shortstack[l]{MAB\\SST}}
 & 20  & 1.358 & 1.279 & 1.132$^\dagger$ & 1.129$^\dagger$ & \textbf{1.115} \\
 & 50  & 1.235 & 1.141 & 1.073$^\dagger$ & 1.035$^\dagger$ & \textbf{1.016} \\
 & 200 & 0.960 & 0.941 & 0.925 & 0.864 & \textbf{0.843} \\
\midrule
\multirow{3}{*}{\shortstack[l]{MAB\\SSS}}
 & 20  & 0.595 & 0.582 & 0.574 & \textbf{0.541} & 0.543$^\dagger$ \\
 & 50  & 0.587 & 0.567 & 0.565 & \textbf{0.497} & 0.500$^\dagger$ \\
 & 200 & 0.516 & 0.507 & 0.516 & 0.411$^\dagger$ & \textbf{0.406} \\
\midrule
\multirow{3}{*}{\shortstack[l]{CCS\\SST}}
 & 20  & 0.664 & 0.620 & \textbf{0.570} & 0.582$^\dagger$ & 0.582$^\dagger$ \\
 & 50  & 0.610 & 0.545 & \textbf{0.466} & 0.494$^\dagger$ & 0.487$^\dagger$ \\
 & 200 & 0.411 & 0.389 & 0.394 & 0.356$^\dagger$ & \textbf{0.351} \\
\midrule
\multirow{3}{*}{\shortstack[l]{CCS\\SSS}}
 & 20  & 0.199 & 0.194 & 0.190 & 0.174$^\dagger$ & \textbf{0.172} \\
 & 50  & 0.194 & 0.182 & 0.162$^\dagger$ & \textbf{0.153} & 0.156$^\dagger$ \\
 & 200 & 0.164 & 0.149 & 0.147 & 0.127$^\dagger$ & \textbf{0.127} \\
\bottomrule
\end{tabular}
\end{table}
}

\textbf{Scoring and reliability.}
We use the RMSE of the ensemble mean to evaluate our method. For each run we compute the
average RMSE of the final eight cycles, as a measure of long-horizon performance
of our closed-loop framework.
Let $S$ denote this score. We compare it with two references.
The \emph{free-running} forecast assimilates no observations, while the
\emph{autoencoder reconstruction baseline} is the error obtained by
encoding and decoding the groundtruth state. We use these measures as an effective
floor and ceiling of the performance, and report $S$ as a fraction of
the gap between these references that is closed,
\(
F=\frac{S^{\mathrm{free}}-S}
        {S^{\mathrm{free}}-S^{\mathrm{recon}}},
\)
so that $F=0$ corresponds to the free-running forecast and $F=1$ to the
autoencoder reconstruction baseline. We average over seeds first and
then over ICs; reported variation is therefore across ICs.

Repeated runs can differ slightly because of numerical nondeterminism, even with the same random seed. To estimate this variation, we rerun the A-optimal experiment for each initial condition using the same seed and define the numerical tolerance as the largest absolute change in score. We use a paired Wilcoxon signed-rank test to check whether one strategy consistently outperforms another across initial conditions, rather than appearing better because of a few individual cases. We consider two strategies meaningfully different when this test gives $p<0.05$ and their mean difference also exceeds the numerical tolerance.

For each initial condition, we also measure how well ensemble uncertainty reflects forecast error. At 240,h, we compute the spatial correlation $r$ between ensemble spread and the absolute error of the ensemble mean across wet cells; higher correlation means that high-uncertainty regions tend to coincide with regions of high forecast error.

\section{Results}
\label{sec:results}

We analyze the experiments to answer the questions:
\begin{enumerate}
    \item[\textbf{Q1.}] Does assimilating the fleet's observations
    reliably improve on the free-running forecast?
    \item[\textbf{Q2.}] How much do mobility and uncertainty-driven
    placement each improve performance?
    \item[\textbf{Q3.}] How does performance scale with fleet size?
    \item[\textbf{Q4.}] When does uncertainty-driven sampling fail
    to outperform simpler strategies, and why?
    \item[\textbf{Q5.}] What is the computational cost of the
    closed-loop system?
\end{enumerate}

\begin{figure}[tb]
\centering
\includegraphics[width=0.99\linewidth]{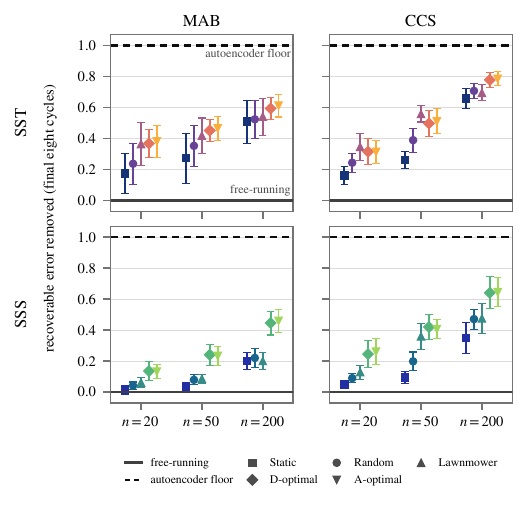}
\caption{\textbf{Closed-loop skill.} Fraction of recoverable error
removed (0 = free-running, 1 = autoencoder floor) for the five strategies
at $n=20/50/200$, by region and variable; mean $\pm 1$ sd across the ten
ICs. The uncertainty-driven pair leads everywhere except CCS temperature
at the two smaller fleets, where \emph{Lawnmower} has the best mean
(Sec.~\ref{sec:q3}).}
\label{fig:summary}
\end{figure}

\begin{figure*}[tp]
\centering
\includegraphics[width=0.99\textwidth]{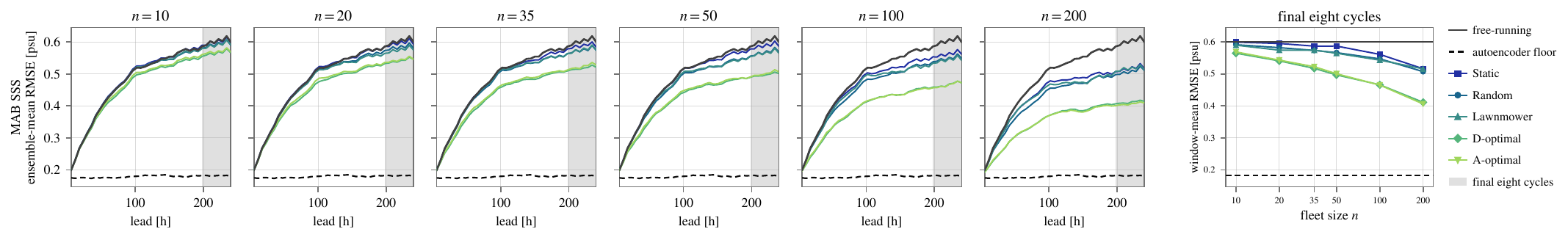}
\caption{\textbf{Effect of fleet size.} MAB salinity at six fleet
sizes. Left, RMSE of the analysis-ensemble mean over the 240\,h run,
averaged over ten ICs and three paired seeds, with the final eight-cycle scoring window
shaded. Right, scoring-window RMSE against fleet size; the horizontal
line shows the free-running score. \emph{A-optimal} remains below \emph{Random}
throughout the sweep, and the gap grows with $n$.}
\label{fig:fleetsize}
\end{figure*}

\begin{figure}
\includegraphics[width=0.99\linewidth]{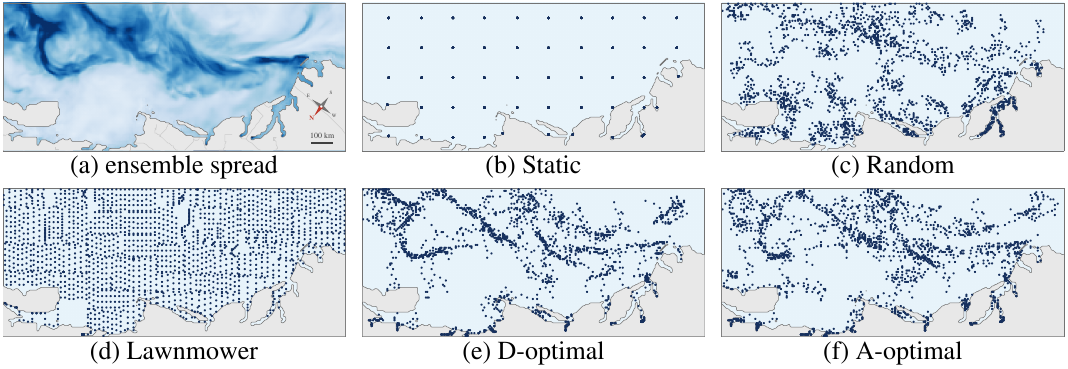}
\caption{\textbf{Where each policy samples.} One MAB IC (2022-04-09,
$n{=}50$, seed 0): the forecast-spread map with every location sampled
over the forty cycles. \emph{Static} remains on its initial network, \emph{Random}
disperses through the reachable domain, \emph{Lawnmower} follows its survey
routes, and \emph{A-optimal} and \emph{D-optimal} concentrate on similar high-spread
structures.}
\label{fig:patterns}
\end{figure}

\begin{figure}[tb]
\centering
\includegraphics[width=0.99\linewidth]{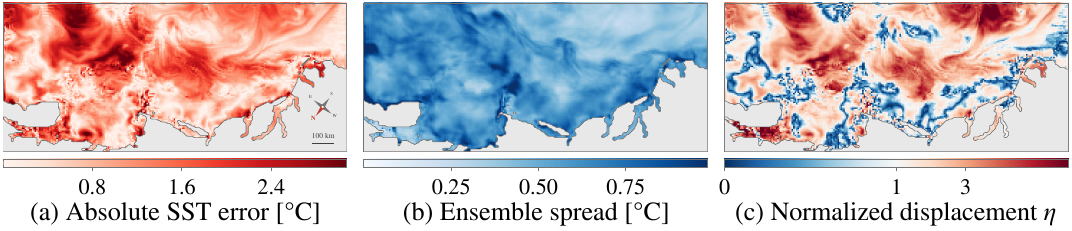}
\caption{\textbf{A confidently wrong belief.} MAB temperature on
2022-08-02 at 240\,h. (a) Absolute error of the ensemble mean.
(b) Ensemble spread. (c) Spread-normalized error $\eta$. Values below $\eta=1$ indicate that the absolute error is
    smaller than one local ensemble standard deviation, while larger values
    mark regions where the forecast mean is far from the truth yet the
    ensemble is confident. The largest
errors occur in regions where the ensemble expresses little
uncertainty.}
\label{fig:diag}
\end{figure}

The overall analysis is driven by the results shown in \cref{tab:summary,tab:peric} and \cref{fig:summary,fig:fleetsize,fig:patterns,fig:diag}.

\cref{fig:show_both}
illustrates one representative IC from each region and highlights the different spatial structure of their forecast errors. In MAB errors tend to concentrate around fronts,
whereas in CCS they are finer-grained and more broadly distributed.

\cref{fig:summary} summarizes the error removed by each sampling
policy across fleet sizes. \cref{tab:summary} reports the
corresponding RMSE over the final eight cycles and marks policies not
reliably separated from the best in each row under the criterion in
\cref{sec:setup}. We use these results to answer the five questions in turn.

\label{sec:q1}
\textbf{(Q1)} Yes, our assimilation step improves the ocean state estimate. Of
the 1,800 runs, 1,787 finish with lower RMSE than the corresponding free-running
forecast, and every policy at every fleet size improves on free-running on
average (\cref{fig:summary}). The thirteen exceptions are all MAB salinity
at $n=20$ or $50$, ten under \emph{Static} and three under \emph{Random}; no \emph{Lawnmower},
\emph{D-optimal}, or \emph{A-optimal} run finishes worse.

\label{sec:q2}
\textbf{(Q2)} See \cref{tab:summary}. Mobility alone, without an intelligent
sampling strategy provides modest benefit over \emph{Static} sensors. The \emph{Static}
strategy consistently had the highest error score of any of the strategies, and
was consistently outperformed by the \emph{Random} and \emph{Lawnmower}
strategies.
Using forecast uncertainty to choose where to move gives the larger and
more consistent gain than randomly sampling. \emph{A-optimal} has lower mean RMSE than \emph{Random} in all
12 region--variable--fleet combination
(\cref{tab:peric}). 
Additionally, there was no substantial difference between the results of
\emph{A-optimal} or \emph{D-optimal}. The two approaches typically obtained
similar scores, matching the classic knowledge in theory of optimal design that
the choice of optimality criterion typically does not substantially affect
the outcomes.

\textbf{(Q3)} Increasing the team size led to improved performance for all methods. The
improvement appeared to be sublinear with respect to team size, which would
indicate a `diminishing returns' effect with increased number of robots due to
overlapping information gained from samples. Both of these observations match
expectations.

\label{sec:q3}
\textbf{(Q4):} Uncertainty-driven sampling loses its advantage when the spatial
uncertainty represented by the forecast ensemble is a poor proxy for forecast
error. We see this limitation in two scenarios. In general, for CCS temperature, the
correlation between ensemble spread and model error is relatively weak across
many initial conditions. For certain instances on the MAB temperature experiments
the ensemble assigns little spread to regions with large forecast error.
We show one example from the MAB cases in \cref{fig:diag}.

In these cases, as the sampling strategies rely on ensemble spread as a proxy
for model error, the mismatch between spread and error limits the performance of
the \emph{A-optimal} and \emph{D-optimal} sampling strategies.
\cref{tab:summary} shows the clearest aggregate exception. On CCS temperature at
$n=20$ and $50$, \emph{Lawnmower} has the lowest mean RMSE, although the
differences are generally small. \emph{Lawnmower} beats both uncertainty-driven
strategies in 15 of 30 runs at $n=20$ and 23 of 30 at $n=50$
(Table~\ref{tab:lawnwins}). This is likely because the coverage-based strategy
of \emph{Lawnmower} increases the odds of sampling areas of high model error,
whereas \emph{A-} and \emph{D-optimal} strategies would only do so if it corresponds
with high ensemble spread.

\label{sec:q4}
\textbf{(Q5)} The full loop is fast relative to the observing
interval, and its cost changes little with fleet size. Every run used
a single NVIDIA A40 in bf16 at batch size one. Across all 1{,}800
runs, the median wall-clock time is 123\,s per cycle on MAB and
151\,s on CCS, within a six-hour observation interval. For \emph{A-optimal},
increasing the fleet from 20 to 200 robots increases cycle time by
only 1.5--2.3\% across the four region--variable pairs, and the
additional cost over \emph{Random} is at most 1.9\% of the full cycle. Most
of the computation is spent updating the belief rather than choosing
where to sample: a free-running cycle takes about 26\,s on MAB and
33\,s on CCS, and in one MAB timing breakdown, guided assimilation
accounts for about 76\% of the full cycle. A complete 40-cycle run ($40 \times 6 = 240$\,h of experiment)
requires compute-hours, on average, of 1.4\,h on MAB and 1.7\,h on CCS.

{
\setlength{\tabcolsep}{3pt}
\renewcommand{\arraystretch}{0.9}
\begin{table}[tb]
\centering\footnotesize
\setlength{\tabcolsep}{2.8pt}
\caption{Per-initial-condition comparison of uncertainty quality and sampling benefit.
$r$ measures how well ensemble spread identifies regions of large forecast error at 240,h; larger values indicate that uncertainty is a better spatial guide to model error. $\Delta_n$ is the RMSE of \emph{Random} minus \emph{A-optimal} for fleet size $n$, so positive values indicate an advantage for uncertainty-driven sampling and negative values favor \emph{Random}. $^\dagger$ marks differences larger than the numerical variation observed when rerunning the same experiment.}
\label{tab:peric}
\begin{tabular}{l r rrr r rrr}
\toprule
 & \multicolumn{4}{c}{SST} & \multicolumn{4}{c}{SSS} \\
\cmidrule(lr){2-5}\cmidrule(lr){6-9}
Date & $r$ & $\Delta_{20}$ & $\Delta_{50}$ & $\Delta_{200}$
     & $r$ & $\Delta_{20}$ & $\Delta_{50}$ & $\Delta_{200}$ \\
\midrule
\multicolumn{9}{l}{\emph{MAB}}\\
2022-01-08 & $+0.59$ & $.183^\dagger$ & $.194^\dagger$ & $.145^\dagger$ & $+0.66$ & $.068^\dagger$ & $.096^\dagger$ & $.159^\dagger$ \\
2022-02-25 & $+0.58$ & $.176^\dagger$ & $.223^\dagger$ & $.200^\dagger$ & $+0.63$ & $.023$ & $.046^\dagger$ & $.115^\dagger$ \\
2022-04-09 & $+0.47$ & $.155$ & $.204^\dagger$ & $.136^\dagger$ & $+0.69$ & $.084^\dagger$ & $.129^\dagger$ & $.162^\dagger$ \\
2022-05-07 & $+0.53$ & $.355^\dagger$ & $.113^\dagger$ & $.111^\dagger$ & $+0.59$ & $.053^\dagger$ & $.088^\dagger$ & $.129^\dagger$ \\
2022-08-02 & $+0.09$ & $.064$ & $.002$ & $-.035^\dagger$ & $+0.56$ & $.020$ & $.032^\dagger$ & $.066^\dagger$ \\
2022-10-14 & $+0.38$ & $.115$ & $.059$ & $.090^\dagger$ & $+0.48$ & $.022$ & $.083^\dagger$ & $.110^\dagger$ \\
2022-11-02 & $+0.51$ & $.256^\dagger$ & $.215^\dagger$ & $.106^\dagger$ & $+0.55$ & $.043^\dagger$ & $.040^\dagger$ & $.067^\dagger$ \\
2022-12-13 & $+0.47$ & $.134$ & $.109^\dagger$ & $.068^\dagger$ & $+0.62$ & $.027^\dagger$ & $.049^\dagger$ & $.046^\dagger$ \\
2023-03-11 & $+0.54$ & $.136$ & $.153^\dagger$ & $.155^\dagger$ & $+0.66$ & $.027^\dagger$ & $.051^\dagger$ & $.090^\dagger$ \\
2023-09-22 & $-0.10$ & $.066$ & $-.021$ & $.004$ & $+0.48$ & $.024^\dagger$ & $.046^\dagger$ & $.068^\dagger$ \\
\midrule
\multicolumn{9}{l}{\emph{CCS}}\\
2016-03-05 & $+0.14$ & $.010$ & $.037$ & $.025^\dagger$ & $+0.43$ & $.021^\dagger$ & $.020^\dagger$ & $.017^\dagger$ \\
2016-04-25 & $+0.21$ & $.008$ & $.027$ & $.035^\dagger$ & $+0.55$ & $.012^\dagger$ & $.021^\dagger$ & $.017^\dagger$ \\
2016-07-27 & $+0.44$ & $.071^\dagger$ & $.159^\dagger$ & $.060^\dagger$ & $+0.63$ & $.069^\dagger$ & $.072^\dagger$ & $.045^\dagger$ \\
2016-08-13 & $+0.25$ & $.022$ & $.004$ & $.047^\dagger$ & $+0.44$ & $.019^\dagger$ & $.025^\dagger$ & $.019^\dagger$ \\
2016-12-03 & $+0.08$ & $.005$ & $.023$ & $.029^\dagger$ & $+0.31$ & $.008$ & $.013^\dagger$ & $.013^\dagger$ \\
2017-01-17 & $+0.01$ & $.020$ & $.064$ & $.032^\dagger$ & $+0.35$ & $.007$ & $.012^\dagger$ & $.010$ \\
2017-06-19 & $+0.48$ & $.074^\dagger$ & $.049$ & $.056^\dagger$ & $+0.45$ & $.027^\dagger$ & $.034^\dagger$ & $.050^\dagger$ \\
2017-09-14 & $+0.31$ & $.098^\dagger$ & $.073^\dagger$ & $.025^\dagger$ & $+0.51$ & $.012^\dagger$ & $.022^\dagger$ & $.018^\dagger$ \\
2017-10-24 & $+0.31$ & $.051^\dagger$ & $.099^\dagger$ & $.043^\dagger$ & $+0.47$ & $.021^\dagger$ & $.024^\dagger$ & $.014^\dagger$ \\
2018-11-21 & $+0.15$ & $.018$ & $.052$ & $.027^\dagger$ & $+0.47$ & $.025^\dagger$ & $.020^\dagger$ & $.014^\dagger$ \\
\bottomrule
\end{tabular}
\end{table}
}

\begin{table}[tb]
\centering
\scriptsize
\setlength{\tabcolsep}{5pt}
\renewcommand{\arraystretch}{0.9}
\caption{\emph{Lawnmower} wins over both uncertainty-driven policies on SST.
Each entry gives wins / above-threshold wins / ICs with at least one
win, out of 30 runs and ten ICs per fleet size.}
\label{tab:lawnwins}
\begin{tabular}{lccc}
\toprule
& $n=20$ & $n=50$ & $n=200$ \\
\midrule
MAB & 7 / 0 / 3  & 6 / 1 / 3  & 3 / 3 / 1 \\
CCS    & 15 / 3 / 7 & 23 / 1 / 8 & 0 / 0 / 0 \\
\bottomrule
\end{tabular}
\end{table}

\textit{Comparison with numerical ocean forecasting.}
For operational context, Wilkin et al.~\cite{wilkin2018coastal}
report timings for a MAB system of the same regional model family and
horizontal resolution (with 40 rather than our 20 vertical levels): on
12 cores of a 3.5\,GHz Intel Xeon system, a three-day 4D-Var analysis
required 155\,min, followed by 15\,min for a deterministic 72\,h
forecast. Over the same 72\,h horizon, our six-hour interval gives 12
cycles: the free-running $N=25$ ensemble requires about
$12\times26\,\mathrm{s}=5.2$\,min on the A40, and 12 complete
forecast--assimilate--plan cycles about
$12\times123\,\mathrm{s}=24.6$\,min. Under the reported hardware configurations, the learned model produces 25 72-h forecast trajectories in about one third of the time ROMS requires for one, while the complete closed-loop sequence takes 24.6,min versus 170,min for the numerical workflow, corresponding to an approximately $7\times$ reduction in wall-clock time.

\section{Limitations and Future Work}
\label{sec:limitations}

We simplified several parts of the problem to keep the study
controlled. We assumed a centralized fleet with perfect
communication, point measurements, and ten days of travel from
deployment. Robots are advected by the smoothed ensemble-mean
current, and their motion is modeled by a simple distance budget
rather than probabilistic reachability
\citep{lolla2014realistic,subramani2018stochastic}. The belief is
also initialized by encoding the reference state, which is not
available in a real deployment.

Our sampling policies are myopic, and their performance depends on
the uncertainty represented by the forecast ensemble. We saw both
limits in Sec.~\ref{sec:q3}: when little of the domain is reachable,
systematic coverage becomes competitive, and when ensemble spread is
poorly aligned with forecast error, uncertainty-driven placement can
lose its advantage. This motivates non-myopic multi-robot planning,
better-calibrated generative forecasts, and sampling criteria that
stay useful when the spread is a poor guide to error. Reducing the
cost of guided assimilation, which dominates the cycle time
(Sec.~\ref{sec:q4}), is also of substantial interest.

\section{Conclusion}
\label{sec:conclusion}

We presented a closed-loop multi-robot adaptive sampling framework in which a learned generative ocean model serves as the predictive belief used for forecasting, uncertainty-aware sampling, robot reachability, and measurement assimilation. We evaluated the framework across two ocean regions, sea-surface temperature and salinity, and fleet sizes of 20, 50, and 200 robots, for 3 seeds, totaling 1,800 ten-day closed-loop experiments. Across these experiments, assimilating robot observations consistently improved the ocean-state estimate, while uncertainty-driven sampling generally outperformed static, random, and systematic-coverage baselines. The low-rank structure of the forecast ensemble also enables efficient evaluation of the sampling objectives, making planning a small fraction of the overall closed-loop computational cost.

Our experiments also expose an important limitation of uncertainty-driven adaptive sampling. When ensemble spread is well aligned with forecast error, the generative belief provides useful guidance about where measurements are most valuable. When the model is confidently wrong, however, regions of large forecast error may receive little sampling priority, and systematic coverage can become competitive or superior. This suggests that the usefulness of generative models for adaptive sensing depends not only on forecast accuracy, but also on the spatial quality of their predictive uncertainty. Future work should therefore combine improved uncertainty calibration with non-myopic multi-robot planning and more realistic models of reachability and communication. More broadly, these results suggest that learned generative forecasts can provide a computationally practical foundation for closed-loop robotic ocean observing systems.

\printbibliography

\end{document}